\documentclass[letterpaper]{article}
\usepackage{aaai}
\usepackage{times}
\usepackage{xcolor}
\usepackage{soul}
\usepackage[utf8]{inputenc}
\usepackage[small]{caption}
\newtheorem{definition}{Definition}
\usepackage{xspace}
\usepackage{algorithm}
\usepackage[noend]{algorithmic}
\usepackage{graphicx} 
\usepackage{url}
\usepackage{multirow,multicol, makecell, booktabs}
\usepackage{amsmath,amssymb}
\usepackage{helvet}
\usepackage{courier}
\usepackage{array}
\usepackage{float}
\usepackage{placeins}
\usepackage{lipsum}

\newcommand{\thickhline}{%
    \noalign {\ifnum 0=`}\fi \hrule height 1pt
    \futurelet \reserved@a \@xhline
}
\begin{document}
%

\newcommand{\alberto}[1]{\textcolor{blue}{#1}}

\newcommand{\daniel}[1]{\textcolor{red}{#1}}
\newcommand{\susana}[1]{\textcolor{violet}{#1}}

\newcommand{\preventing}{\textsc{prev}\xspace}
\newcommand{\seeking}{\textsc{seek}\xspace}
\newcommand{\algoritmocp}{\textsc{dicp}\xspace}
\newcommand{\algoritmoacp}{\textsc{adicp}\xspace}

\renewcommand{\algorithmicrequire}{\textbf{Input:}}
\renewcommand{\algorithmicensure}{\textbf{Output:}}
\nocopyright

\title{Anticipatory Counterplanning}
\author{Alberto Pozanco$^1$, Yolanda E-Martín$^2$, Susana Fernández$^1$, Daniel Borrajo$^1$\\
$^1$Departamento de Informática, Universidad Carlos III de Madrid\\
$^2$Florida Universitària\\
alberto.pozanco@gmail.com, yescudero@florida-uni.es, sfarregu@inf.uc3m.es, dborrajo@ia.uc3m.es
}

\maketitle
\begin{abstract}
\begin{quote}
In competitive environments, commonly agents try to prevent opponents from achieving their goals. 
Most previous preventing approaches assume the opponent's goal is known a priori. Others only start executing actions once the opponent's goal has been inferred.
In this work we introduce a novel domain-independent algorithm called Anticipatory Counterplanning. It combines inference of opponent's goals with computation of planning centroids to yield proactive counter strategies in problems where the opponent's goal is unknown.
Experimental results show how this novel technique outperforms reactive counterplanning, increasing the chances of stopping the opponent from achieving its goals.
\end{quote}
\end{abstract}

\section{Introduction}
In competitive environments, commonly agents try to prevent opponents from achieving their goals, as in security~\cite{DBLP:conf/aips/BoddyGHH05,DBLP:conf/atal/PitaJMOPTWPK08}, real-time strategy games~\cite{DBLP:journals/tciaig/OntanonSURCP13}, or air combat~\cite{DBLP:conf/flairs/BorckKAA15}.
There exist different approaches to frame and solve these kind of problems~\cite{DBLP:journals/ai/Carbonell81,DBLP:conf/iaw/Rowe03,DBLP:books/daglib/0040483,DBLP:conf/aaai/SpeicherS00K18}.
However, most of these works assume the opponent's goal is known a priori, which is not true in many real-world problems.

\begin{figure}
    \centering
    \includegraphics[scale=0.45]{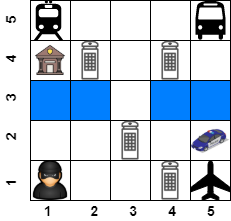}
    \caption[Police control domain]{Police control domain where the police  wants to stop a terrorist (depicted by a masked agent) that is trying to leave the city after
     committing an attack. Blue tiles depict a river that agents cannot traverse. Transport icons in the corners depict different stations that the terrorist can use to escape. The building icon depicts the police station, and the car a police patrol. Phone booths are depicted by black and white cabin icons.}
    \label{fig:intro}
\end{figure}

For example, consider a police control domain like the one shown in Figure~\ref{fig:intro}, where two agents act concurrently in the same environment. A terrorist has committed an attack and
wants to escape, while the police aim at stopping the terrorist before she leaves the city.  The terrorist has carefully
designed her escape plan buying bus tickets; she therefore needs to go to the bus station and take the bus. 
Before that, she needs to make a call to a partner.
However, she is afraid that her phone is tapped by the police, so she needs to make the call from any of the phone booths distributed over the city.
Once she
reaches the bus station having made the call from a non-tapped phone, she will be out of reach of the police. 
On the other side, the police do not know which means of
transport from a given known set (e.g. bus, train or plane) the terrorist is going to use to escape. 
The police can move around the city using a patrol car and set controls in the white tiles (blue tiles are obstacles, as a river).
They can also tap the phone booths from the police station.
The police have control over some of the city
cameras, located at different key points around the city, which helps them to identify the terrorist's executed actions.
If we would like to use most of the existing techniques, we would need to group all the possible terrorist's goals into one, thus rendering the problem unsolvable, i.e., no strategy would allow the police to prevent the terrorist from achieving \emph{any} of the goals.

A different approach called domain-independent Counterplanning~\cite{DBLP:conf/ijcai/PozancoEFB18} was recently proposed to prevent opponents from achieving their goal in scenarios like the above, where that goal is unknown a priori.
Counterplanning combines different techniques to generate \emph{counterplans}: (1) planning-based goal recognition~\cite{DBLP:conf/ijcai/RamirezG09} to identify the actual opponent's goal from a predefined set of candidate goals; (2) landmarks~\cite{DBLP:journals/jair/HoffmannPS04} to identify subgoals the opponent will need to achieve to reach its goal; and (3) classical automated planning to generate plans that delete these subgoals before the opponent stops needing them.

However, Pozanco et al. work presents a key drawback: the agent stands still observing the opponent until it infers the opponent's goal, at which point it starts computing and executing a counterplan.
This passiveness (reactiveness) of the agent renders many problems to be unsolvable, since at that point the opponent might be \emph{closer} to her goal, and therefore no counterplan could prevent it from achieving the goal.
This is the case of the scenario depicted in Figure~\ref{fig:intro}, where current counterplanning techniques could not produce any valid counterplan. Another limitation of previous works was that they were generating plans in an off-line manner.

This paper overcomes these problems by making the agent  proactive in on-line settings.
A proactive agent would start executing actions that would allow her to be \emph{in a better spot} for stopping the opponent once it infers her goal.
In the case of Figure~\ref{fig:intro}, the police patrol should start moving west regardless of the opponent's first action. Thus, they would be closer on average to the potential opponent's goals, therefore being closer to stop it.
We argue that planning centroids~\cite{DBLP:conf/aips/PozancoEFB19,karpasCentroids} are these \emph{better spots}, being the states towards the agent should start moving prior to inferring the opponent's goal.

The main contributions of this paper are: (1) adaptation of previous counterplanning techniques to an on-line setting; and (2) definition of anticipatory strategies for counterplanning. Experimental results on several domains show the benefits of using anticipation over reactive behavior.

\section{Background}
\subsection{Automated Planning}
Automated Planning is the task of choosing and organizing a sequence of actions such that, when applied in a given
initial state, it results in a goal state~\cite{DBLP:books/daglib/0014222}. Formally:
\begin{definition}
  A \textbf{single-agent} {\sc strips} \textbf{planning task}
can be defined as a tuple $\Pi=\langle F,A,I,G\rangle $, where $F$ is a set of propositions, $A$ is a set of
instantiated actions, $I\subseteq F$ is an initial state, and $G\subseteq F$ is a set of goals.  
\end{definition}

A state consists of a set of propositions $s\subseteq F$ that are true at a given time. A state is totally specified if
it assigns truth values to all the propositions in $F$, as the initial state $I$ of a planning task. A state is
partially specified (partial state) if it assigns truth values to only a subset of the propositions in $F$, as the
conjunction of propositions $G$ of a planning task.
Each action $a\in A$ is
described by a set of preconditions (pre($a$)), which represent literals that must be true in a state to execute an
action, and a set of effects (eff($a$)), which are the literals that are added (add($a$) effects) or removed (del($a$)
effects) from the state after the action execution. The definition of each action might also include a cost $c(a)$ (the
default cost is one). The execution of an action $a$ in a state $s$ is defined by a function $\gamma$ such that
$\gamma(s,a)=(s\setminus\mbox{del}(a))\cup\mbox{add}(a)$ if pre($a$)$\subseteq s$, and $s$ otherwise (it cannot be
applied). The output of a planning task is a sequence of actions, called a plan, $\pi=(a_1,\ldots,a_n)$. The execution
of a plan $\pi$ in a state $s$ can be defined as:

\begin{small}
\[\Gamma(s,\pi)=\left\{\begin{array}{ll}
                        \Gamma(\gamma(s,a_1),(a_2,\ldots,a_n)) & \mbox{if } \pi\neq \emptyset\\
                        s & \mbox{if } \pi=\emptyset\\
                      \end{array}
                    \right.
\]
\end{small}

A plan $\pi$ is valid if $G\subseteq\Gamma(I,\pi)$. The plan cost is commonly defined as
$c(\pi)=\sum_{a_i\in\pi} c(a_i)$. 
A plan with minimal cost is called optimal.
We will use $h^*(s,G)$ to denote the optimal cost of reaching a goal state $G$ from a given state $s$ of a planning task.
Finally, we will use the function {\sc Planner}($\Pi$) to refer to an algorithm that computes a plan $\pi$ from a planning task $\Pi$.

When several agents act in the same environment, it becomes non-deterministic due to 
agents not knowing in advance which actions the other agents are executing.
Classical plans are no longer valid solutions for these problems, and we need to extend these definitions.
Following Bowling, Jensen and Veloso (\citeyear{DBLP:conf/ijcai/BowlingJV03}), and  Brafman and Domshlak
(\citeyear{DBLP:conf/aips/BrafmanD08}) formulations:
\begin{definition}
  A \textbf{multi-agent planning task} is a tuple $\mbox{MAP} = \langle N, F_i,A_i,I_i,G_i\rangle$, where $N = \{1,\ldots,n\}$ is a set of agents,
 $F_i$ is the set of propositions of agent $i \in N$,
 $I_i$ is the initial state of agent $i \in N$,
 $A_i$ is the set of actions agent $i \in N$ can execute, and
$G_i$ is the goal for agent $i \in N$.
\end{definition}

The set of actions of each agent always includes a no-op action ($\emptyset$).
A solution to a MAP task will be $n$ plans that will be jointly executed. 
To ensure that joint actions have well-defined effects, it is necessary to impose concurrency constraints that model whether a set of actions can be performed in parallel. 
We will assume the propositions-based concurrency constraints introduced in PDDL 2.1~\cite{DBLP:journals/jair/FoxL03}, and only two agents executing one action at each time step.
Two actions $a_1$ and $a_2$ can
be applied concurrently in a state $s$ only iff $\mbox{pre}(a_1) \cup \mbox{pre} (a_2) \subseteq s$, and they do not interfere~\cite{DBLP:journals/jair/FoxL03}.
We will use $\gamma_J$ to represent the joint execution of two actions.

\begin{figure*}
\small
\begin{equation}
\label{eq1}
\gamma_J(s,a_1,a_2)=\left\{\begin{array}{ll}
                         (s\setminus (del(a_1)\cup del(a_2)) \cup add(a_1) \cup
add(a_2) &   if\, a_1\,  \mbox{and}\, a_2\, \mbox{do not interfere}\\
                         s &  if\, a_1\, \mbox{and}\, a_2\, \mbox{interfere}\\
                         \gamma(s,a_1) &  if\, a_2\, =\, \emptyset\\
                         \gamma(s,a_2) &  if\, a_1\, =\, \emptyset\\
                      \end{array}
                    \right.
\end{equation}


\begin{equation}
\label{eq2}
 \Gamma_J(s,\pi_j, \pi_k)=\left\{\begin{array}{ll}
                         \Gamma_J(\gamma_J(s,a_{j,1},a_{k,1})),(a_{j,2},a_{k,2}),\ldots,(a_{j,n},a_{k,n})) &  if\, \pi_j , \pi_k \neq \emptyset\\
                         \Gamma(\gamma(s,a_{k,1}),(a_{k,2},\ldots a_{k,n})) &  if\, \pi_j=\emptyset\\
                         \Gamma(\gamma(s,a_{j,1}),(a_{j,2},\ldots a_{j,n})) &  if\, \pi_k=\emptyset\\
                      \end{array}
                    \right.
\end{equation}

\end{figure*}

\begin{definition}
  The \textbf{joint execution of two actions} $a_1,a_2$ in a state $s$ results in a new state given by Equation~\ref{eq1}. 
\label{def:joint_action}
\end{definition}

Similarly, we define the joint execution of two plans as the iterative joint execution of the actions of those plans.

\begin{definition}
  The \textbf{joint execution of two plans} $\pi_j,\pi_k$ in a state $s$ results in a new state given by Equation~\ref{eq2}. 

\end{definition}

Following Cimatti et al. terminology (\citeyear{DBLP:journals/ai/CimattiPRT03}) we define a strong plan in the context of two agents MAP as follows:
\begin{definition}
  A plan $\pi_1$ that solves an agent's planning task $\Pi_1$ is a \textbf{strong plan} iff its joint execution with any sequence of actions $\pi_2$ that an agent with planning task $\Pi_2$ can execute always achieves the goal $G_1 \in \Pi_1$:
  \begin{equation}
      \forall \pi_k \in \Pi_2, G_1 \subseteq \Gamma_J(I_1,\pi_1,\pi_k)
  \end{equation}
\end{definition}
Any other plan that does not meet this criteria will be weak.

\subsection{Goal Recognition}

Goal Recognition is the task of inferring another agent' goals through the observation of its interactions with the
environment. The problem has captured the attention of several computer science
communities~\cite{DBLP:journals/ai/AlbrechtS18}. Among them, planning-based goal
recognition approaches have been shown to be a valid domain-independent alternative to infer agents'
goals~\cite{DBLP:conf/ijcai/RamirezG09,DBLP:conf/aaai/KaminkaVA18,DBLP:journals/ai/PereiraOM20}.
Ramírez and Geffner~(\citeyear{DBLP:conf/aaai/RamirezG10}) developed an approach that assumes observations are actions,
and formally defined a planning-based goal recognition problem as:

\begin{definition} A \textbf{goal recognition problem} is a tuple $T=\langle P,{\cal G},O, Pr\rangle $
  where $P=\langle F,A,I\rangle $ is a planning domain and initial conditions, $\cal G$ is the set of possible goals
  $G$, $G \subseteq F$, $O=(o_1,....,o_m)$ is an observation sequence with each $o_i$ being an action in $A$, and $Pr$
  is a prior probability distribution over the goals in $\cal G$.
\end{definition}

The solution to a goal recognition problem is a probability distribution over the set of goals $G \in {\cal G}$ giving
the relative likelihood of each goal. In this work we assume that $Pr$ is uniform. We use {\sc Recognize}($F,A,I,{\cal G},O$) to refer to an algorithm that solves the goal recognition problem. 
This function returns the set of goals ${\cal G}^\prime \subseteq {\cal G}$ that are most probable according to the set of observations ${\cal O}$.

\subsection{Planning Centroids}
Planning centroids~\cite{DBLP:conf/aips/PozancoEFB19,karpasCentroids} are the states that minimize the average distance (cost) to a given set of possible goals.
The setting is similar to the single-agent \textsc{strips} planning task with multiple goals, like in a goal recognition problem, and a weight associated to each of the goals, i.e., $\Pi=\langle F,A,I,{\cal G},{W_{\cal G}}\rangle$.
The weighted average cost from a state $s$ to the possible goals ${\cal G}$ is computed as:
\begin{equation}
  \frac{1}{|{\cal G}|}\sum_{G \in {\cal G}, w_G \in W_{\cal G}} w_G \times h^*(s,G)  
\end{equation}
where $w_G$ is a real number denoting the weight or importance of the goal $G \in {\cal G}$.
We adhere to the previous definition of planning centroids.

\begin{definition}
  A state $s$ is a \textbf{centroid state} of a task $\Pi=\langle F,A,I,{\cal G}\rangle$ iff (1) it is reachable from the initial state; and (2) it minimizes the weighted average cost to the set of possible goals ${\cal G}$. 
\end{definition}

Planning centroids can be computed exactly or approximately using different techniques~\cite{DBLP:conf/aips/PozancoEFB19,karpasCentroids}.
We will use the function \textsc{ExtractCentroids}$(F,A,I,{\cal G},W_{\cal G})$ to refer to an algorithm that computes a set of centroids from such planning task.

\subsection{Landmarks}

In Automated Planning, landmarks were initially defined as sets of propositions that have to be true at some time in
every solution plan~\cite{DBLP:journals/jair/HoffmannPS04}. 
Although computing all the landmarks of a planning task has been proven to be equivalent to solving the planning task~\cite{DBLP:journals/jair/HoffmannPS04}, efficient techniques has been developed to find a large number of landmarks in reasonable time~\cite{DBLP:conf/aaai/RichterHW08,DBLP:conf/ecai/KeyderRH10}.
We use  {\sc ExtractLandmarks}$(F,A,I,G)$ to refer to an algorithm that computes a set of landmarks from a planning task $\Pi$.

\section{Counterplanning}
In this section we describe the counterplanning setting we assume.
This setting is similar to the one used in previous works~\cite{DBLP:conf/ijcai/PozancoEFB18}, and we will highlight any difference throughout the section.

\begin{itemize}
    \item We consider two planning agents acting concurrently in the same environment. A seeking agent, \seeking, which wants to achieve a goal; and a preventing agent, \preventing, which wants to stop the seeker from achieving its goal.
    
    \item \seeking will try to achieve a goal $G_{\seeking}$ through a rational (optimal or 'least suboptimal') plan $\pi_\seeking$. Goal and plan will not change during the counterplanning episode.

    \item \preventing's goal is initially set to empty; hence, it does not have an initial plan. She will try to formulate a goal and a plan during the counterplanning episode to prevent \seeking from achieving $G_{\seeking}$. 
    
    \item \preventing knows \seeking's model, but not \seeking's actual goal $G_{\seeking}$ or plan $\pi_{\seeking}$.
    
    \item \preventing and \seeking models are coupled; i.e., they share some propositions ${p \in F}$.
    More specifically, \preventing can delete(add) some propositions appearing in \seeking's actions preconditions.
    
    \item \preventing knows a set of potential goals that \seeking might try to achieve, ${\cal G}_{\seeking}$. \seeking actual goal is always within that set, $G_{\seeking} \subseteq {\cal G}_{\seeking}$.
    
    \item Both agents have full observability of other's actions.
    
    \item Both agents only execute one action at each time step, and the duration of each action is one time step.

\end{itemize}

Most of these assumptions are common either in goal recognition research or real world applications. For instance, in
most real world domains where counterplanning can be useful (e.g. police control, cyber security, strategy
games,\ldots), the preventing agent knows her enemy's model and a set of potential goals that she is interested in.  The
rationality assumption is common in goal recognition
research~\cite{DBLP:conf/ijcai/MastersS17}.
Considering the above assumptions, we can formally define a counterplanning task.

\begin{definition}
A \textbf{counterplanning task} is defined by a tuple\\
  ${\cal C} = \langle \Pi_{\seeking}, \Pi_{\preventing}, {\cal O}_{\preventing}, {\cal G}_{\seeking}\rangle$ where:
\begin{itemize}
\item$\Pi_{\seeking} = \langle F_{\seeking},A_{\seeking}, I_{\seeking}, G_{\seeking} \rangle$ is the planning task of
  $\seeking$. 
\item $\Pi_{\preventing} = \langle F_{\preventing},A_{\preventing}, I_{\preventing}, G_{\preventing} \rangle$ is the
  planning task of $\preventing$.
    
    \item ${\cal O}_{\preventing}=(o_1,\ldots,o_m)$ is a set of observations in the form of executed actions that \preventing receives from the execution of
      \seeking's plan\\
      $\pi_{\seeking}=(o_1,\ldots,o_m,a_{m+1},\ldots,a_k)$. The notation differentiates between
      observations (previously executed \seeking's actions), $o_i$, and future actions to be executed by \seeking,
      $a_j$. 
    
    \item ${\cal G}_{\seeking}$ is the set of goals that \preventing currently thinks \seeking can be potentially 
      pursuing. 
     
\end{itemize}
  \label{counterplanning_task}
\end{definition}

The meaning of \emph{currently} in the definition of ${\cal G}_{\seeking}$ indicates that this set changes according to
the set of observations ${\cal O}_{\preventing}$.
In fact, given that we are assuming rational agent behavior to achieve its goal, the size of ${\cal G}_{\seeking}$ will tend to decrease with each
observation $o \in {\cal O}_{\preventing}$. In other words, \preventing will consider less (or equal) \seeking's potential goals as
\seeking executes more actions of her plan. 

As we have discussed, at the beginning of the counterplanning task \preventing has not performed any action (her goal and plan are empty).
Therefore, the new composite state $I_c$ after
receiving the set of observations ${\cal O}_{\preventing}$ is defined as
$I_c = \Gamma (I_{\seeking} \cup I_{\preventing},{\cal O}_{\preventing})$.
The solution to a counterplanning task is a preventing agent's plan $\pi_{\preventing}$, namely a counterplan.
We define valid counterplans\footnote{From now on we will use the terms counterplan and valid counterplans indistinctly.} as follows:
\begin{definition}
  \label{def:counterplan}
  A counterplan $\pi_\preventing$ is a \textbf{valid counterplan} iff its joint execution from $I_c$ with the remaining of \seeking's actual plan $\pi_\seeking$, results in a state $I^\prime_c$ from which \seeking cannot achieve any of the goals in ${\cal G}_{\seeking}$, and therefore its actual goal $G_{\seeking}$. Formally:
  \begin{equation}
      I^\prime_c = \Gamma_J(I_c,\pi_\preventing,\pi_\seeking \setminus \{ {\cal O}_\preventing \})
  \end{equation}
  \begin{equation}
       \forall G_i \in {\cal G}_\seeking, \nexists \pi^\prime_\seeking \mid G_i \subseteq \Gamma(I^\prime_c,\pi^\prime_\seeking)
  \end{equation}
\end{definition}

Note that the definition of a valid counterplan is quite strict: it must prevent \seeking from achieving \emph{any} of the goals
\preventing thinks she is \emph{currently} trying to achieve. 
Moreover, it will only be a valid counterplan with respect to the \emph{actual} plan \seeking is executing $\pi_\seeking$, which \preventing does not know.
Therefore, the validity of a counterplan can only be tested \emph{a posteriori}.
Going back to our running example, in the limit case where the terrorist has not started moving, ${\cal O}_{\preventing}=\emptyset$, the police would need to compute a counterplan that blocks the achievement of any of the terrorist's goals.
In case such a counterplan does not exist, the police should wait until they infer the terrorist's true goal by observing more actions.
Other approaches would involve \emph{betting} for one of the goals and setting a control at one of the stations. 
However, we are aiming at domains such as police control where we want to provide some guarantees about the opponent being stopped.

\subsection{Counterplanning Landmarks}
In automated planning, the only way of ensuring that a goal is not achievable is to thwart any of the planning landmarks
involved in it (as a reminder, goals are landmarks by definition). 
If those landmarks cannot be achieved again as we are assuming here, 
this would prevent \seeking from achieving the goal regardless the plan it follows.
\preventing does not know \seeking's actual goal but a set of potential goals she might be trying to
achieve ${\cal G}_{\seeking}$.  
\begin{definition}
  Given a counterplanning task ${\cal C}$, we refer to the \textbf{set of all the potential planning tasks} that \preventing currently thinks \seeking might be solving as ${\cal X}_{\seeking}$. 
  \begin{equation}
    {\cal X}_{\seeking} =\{ \langle F_{\seeking}, A_{\seeking}, I_{\seeking},G_{i}\rangle: G_i \in {\cal G}_{\seeking}\}
\end{equation}
\end{definition}

Therefore, \preventing must find a counterplan that deletes (or adds) any of the fact landmarks that are \emph{common} in all the planning tasks in
${\cal X}_{\seeking}$. 
We refer to this set of landmarks as counterplanning landmarks.

\begin{definition}
\label{counterplanning_landmarks}
  Given a counterplanning task $\cal C$, a fact in $L_i \in F_{\seeking}$ is a \textbf{counterplanning landmark} iff:
  \begin{itemize}
      \item $L_i \in {\cal L}_{\Pi_j}, \forall$  $\Pi_j \in {\cal X}_{\seeking}$; 
     
      \item If $L_i$ is a positive literal, $\exists a \in A_{\preventing}$ such that $L_i \in \mbox{del}(a)$. If $L_i$ is a negative literal, $\exists a \in A_{\preventing}$ such that $L_i \in \mbox{add}(a)$; and

      \item \preventing can delete (add) $L_i$ applying less actions from $I_c$ than the last step of an optimal plan in which \seeking needs $L_i$. Given ${\cal P}^*_{{\cal X}_\seeking}$, which contains all the optimal plans that achieve any of the goals $G_i \in {\cal G}$; and a function \textsc{laststep}($L_i$,$\pi$) that returns the last step in which $L_i$ appears in any precondition of a plan $\pi$:
      \begin{equation}
          \scriptstyle \nexists \pi_\seeking \in {\cal P}^*_{\Pi_\seeking}, \textsc{laststep} (L_i,\pi_\seeking) < c(\textsc{Planner}(F_\preventing,A_\preventing,I_c,\neg L_i))
      \end{equation}
  \end{itemize}
\end{definition}

This definition of counterplanning landmark is different from the one used by Pozanco et al. (\citeyear{DBLP:conf/ijcai/PozancoEFB18}).
In their work, a counterplanning landmark can be deleted by \preventing in less steps than \seeking can achieve it. 
However, that definition artificially restricts the number of counterplanning landmarks.
For example, landmarks that are true in the initial state would have a cost of zero for \seeking, and therefore could not be considered as counterplanning landmarks.
We propose a different definition that allows us to correctly compute all the counterplanning landmarks of a counterplanning task.
We compute all the optimal plans that solve each of the planning tasks in ${\cal X}_\seeking$, and extract the minimum last step amongst all the optimal plans in ${\cal X}_\seeking$ at which \seeking stops needing $L_i$.
This is needed in order to keep some stopping guarantees, considering a worst case \seeking agent following the plan that stops needing the landmarks as soon as possible.
We refer to the set of counterplanning landmarks $cpl_i$ of a counterplanning task as ${\mbox{CPL}}$, and
\textsc{ExtractCPL}($\cal C$) as the function that computes them. It returns tuples of the form $\langle cpl^i, c(\pi_\preventing^i) \rangle$, where $\pi_\preventing^i$ is the cost of a plan that solves $\Pi_\preventing^i = \langle F_\preventing, I_c, A_\preventing, cpl^i\rangle$.
As we will see next, 
\preventing will set any of these counterplanning landmarks as her goal $G_{\preventing}$, computing a counterplan that deletes (adds) it, making impossible for \seeking to achieve her goal $G_{\seeking}$.

\subsection{Computing Counterplans}
We now adapt Pozanco \textit{et al.} (\citeyear{DBLP:conf/ijcai/PozancoEFB18}) Domain-independent Counterplanning (\algoritmocp) algorithm to an online setting.
\algoritmocp is shown in Algorithm~\ref{algoritmo_nuevo} with black lettering, while the teal lettering corresponds to the new components of the anticipatory counterplanning algorithm we will discuss later.

\begin{algorithm}
\caption{\textcolor{teal}{\textsc{a}}\algoritmocp}
\label{algoritmo_nuevo}

\small
\begin{algorithmic}[1]

\REQUIRE $\cal C$
\ENSURE $\pi_{\preventing}$

\STATE $\pi_{\preventing}, \pi^\prime_{\preventing}, a_\preventing, a_{\seeking}, G_{\preventing}, {\cal O}_\preventing \gets \emptyset$
\STATE $I_c = I_{\seeking} \cup I_{\preventing}$

\WHILE{$\pi_\seeking \neq \emptyset$ \textbf{and} $\pi_\preventing = \emptyset$}
    \IF{$\pi^\prime_\seeking \neq \emptyset$ \textbf{or} ${\cal O}_\preventing = \emptyset$}
    
    \STATE ${\cal G}_{\seeking} \gets \textsc{Recognize}(F_{\seeking},A_{\seeking},I_{\seeking},{\cal G}_{\seeking},{\cal O}_{\preventing})$
    \STATE $\mbox{CPL} \gets \textsc{ExtractCPL}({\cal C})$
    \STATE \textcolor{teal}{$\mbox{CPList} \gets \textsc{ExtractListOfCPL}({\cal C})$}

    \WHILE{$\pi_{\preventing} = \emptyset$ \textbf{and} $\mbox{CPL} \neq \emptyset$}
        \STATE $G_{\preventing} \gets \textsc{SelectGoal}(\mbox{CPL})$ 
        \STATE $\pi_{\preventing} \gets \textsc{Planner}(F_{\preventing},A_{\preventing},I_c,G_{\preventing})$
        \IF{$\pi_\preventing = \emptyset$}
            \STATE $\mbox{CPL} \gets \mbox{CPL} \setminus G_{\preventing}$
        \ENDIF
    \ENDWHILE

    \IF{$\pi_\preventing = \emptyset$}
    \STATE \textcolor{teal}{$a_{\preventing} \gets \mbox{\textsc{anticipate}}(\Pi_\preventing,\mbox{CPList},I_c)$}
    \STATE $\pi^\prime_\preventing.\textsc{add}(a_\preventing)$
     \STATE $a_\seeking \gets \pi_{\seeking}.\textsc{pop()}$
    \STATE $I_c = \gamma_J (I_c,a_{\seeking},a_\preventing)$
    \STATE ${\cal O}_{\preventing} \gets {\cal O}_{\preventing}.\textsc{insert}(\pi^\prime_\seeking)$
    \ENDIF
    
\ENDIF
\ENDWHILE
\STATE $\pi_\preventing = \pi^\prime_\preventing + \pi_\preventing$
\RETURN $\pi_{\preventing}$

\end{algorithmic}
\end{algorithm}

The algorithm receives a counterplanning task ${\cal C}$ as input and returns preventing's plan $\pi_\preventing$ as output.
The algorithm loops until no more observations are received, i.e., \seeking has reached its goal, or \preventing finds a counterplan to block \seeking.
It first calls the \textsc{Recognize} function (line 5).
Given a planning domain, initial conditions, a set of candidate goals, and a set of observations, this function updates the set of candidate goals ${\cal G}_\seeking$.
After that, it extracts the set of counterplanning landmarks (line 6, see Definition~\ref{counterplanning_landmarks}).
If this set is not empty, a counterplan might exist and it proceeds to find it.
Otherwise, the counterplanning task is unsolvable, i.e., no counterplan exists given the current observations, and we advance the simulator, i.e., update the composite state $I_c = \gamma_J (I_c,\pi^\prime_{\seeking},\pi^\prime_\preventing)$.
This is done in lines 16-18, where it perceives the next \seeking's action $\pi^\prime_\seeking$, inserts it to the set of observations received by \preventing ${\cal O}_\preventing$, and updates the composite state $I_c$.

If the set of counterplanning landmarks is not empty, it first selects a goal $G_\preventing$ from CPL using the \textsc{SelectGoal}.
As discussed by Pozanco et al. (\citeyear{DBLP:conf/ijcai/PozancoEFB18}), there exist different ways of selecting this counterplanning landmark.
For example, it could return the counterplanning landmark that is closer to \preventing, i.e., the one it can achieve with the lowest cost; or closer to \seeking, therefore stopping it as soon as possible.
Finally, it uses a planner to compute a counterplan that achieves $G_\preventing$.
Again, there exist different ways of generating the counterplan.
For example, we could use a planner that only returns strong counterplans, i.e., counterplans that would guarantee the opponent is blocked; or optimal counterplans, that would guarantee minimal cost on the \preventing's plan.
If such (counter)plan exists, the algorithm will return it, concatenating it to the actions executed before by the preventing agent, $\pi^\prime_\preventing$.
Otherwise, it removes that counterplanning landmark from CPL (line 12) and tries to select a new one to set it as $G_\preventing$.
This is done until $CPL=\emptyset$, in which case it advances the simulator (lines 16-18). 

\section{Anticipatory Counterplanning}
Many counterplanning tasks are not solvable
given that \seeking is closer to all the landmarks involved in ${\cal G}_{\seeking}$ than \preventing (CPL$=\emptyset$).
If this happens when \seeking has not even moved, i.e., a counterplanning task with ${\cal O}_{\preventing}=\emptyset$, there is little \preventing can do to block its opponent.
However, in some counterplanning episodes this \preventing's \emph{handicap} comes from the fact that \preventing stands still observing \seeking actions.
Even if \preventing was able to stop \seeking at the beginning of the counterplanning episode (with ${\cal O}_{\preventing}=\emptyset$) now it cannot, since \seeking is now closer to all the landmarks.

\begin{figure}[hbtp]
    \centering
    \includegraphics[width=\columnwidth]{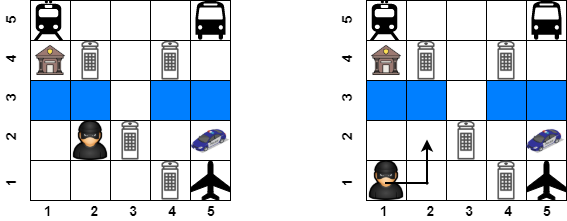}
    \caption[Unsolvable counterplanning tasks \textsc{police control} ]{Unsolvable counterplanning tasks in the \textsc{police control} domain. The task on the left is unsolvable regardless \preventing actions. The task on the right would be solvable if \preventing had started moving at the same time as \seeking in the right direction.}
    \label{fig:anticipatory1}
\end{figure}

We can see these two cases in Figure~\ref{fig:anticipatory1}, which depicts two different counterplanning tasks using our police control running example.
The counterplanning task depicted in the left figure is unsolvable regardless of \preventing actions.
The opponent has not executed any action, and there is no counterplanning landmark that \preventing can falsify before \seeking stops needing it.
On the other hand, the counterplanning task on the right would have a solution if the agent had started moving in the \emph{right} direction at the same time as \seeking executed its first action, instead of standing still and watching.

Previous counterplanning works are \emph{reactive}, assuming \preventing only generates a (counter)plan when it has inferred \seeking's goal.
We can see that in lines 15-18 of Algorithm~\ref{algoritmo_nuevo}, where no action is executed by \preventing ($\pi^\prime_\preventing=\emptyset$) in case there are no common counterplanning landmarks among the most likely goals.
In this work we want to go further, making \preventing \emph{proactive}, i.e., it will start executing actions before inferring \seeking's goal in a way that increases its chances of blocking the opponent once its goal has been inferred.
The key insight is that we can use previous work on computing centroids to guide $\preventing$ towards a reasonable state when it lacks any direction to move according to \algoritmocp.
Hence,
we have modified \algoritmocp to use planning centroids to drive \preventing's actions until a counterplan is found.
We call this new algorithm \algoritmoacp (Anticipatory Domain-independent Counterplanning), and the two additions needed in Algorithm~\ref{algoritmo_nuevo} are represented with teal lettering.

The first addition is line 7, where it calls a new function \textsc{ExtractListOfCPL}.
This function computes the \emph{individual} counterplanning landmarks, i.e., the counterplanning landmarks of each planning task $\Pi \in {\cal X}_\seeking$, rather than only the common ones, i.e., those that appear in all $\Pi \in {\cal X}_\seeking$. 
The second addition is line 14, where it calls a new function  \textsc{anticipate}.
This will make \preventing execute the action prescribed by the new \textsc{anticipate} function until a counterplan $\pi_\preventing$ is found.
\textsc{anticipate} returns the next action to be executed by \preventing based on the computation of the centroid of all the counterplanning landmarks. 
Algorithm~\ref{alg_grscp} details how \textsc{anticipate} works.
\begin{algorithm}
\caption{\textsc{anticipate}}
\label{alg_grscp}
\small
\begin{algorithmic}[1]

\REQUIRE $\Pi_\preventing$,$\mbox{CPList}$,$I_c$
\ENSURE $\pi_{\preventing}$

\STATE $\pi_{\preventing} \gets \emptyset$
\STATE ${\cal G} \gets $\textsc{Rank}(CPList)
\STATE $\mbox{centroid} \gets \mbox{\textsc{ExtractCentroids}} (F_\preventing,A_\preventing,I_c,{\cal G})$
\STATE $a_\preventing \gets \mbox{\textsc{GetFirstAction}}(F_\preventing,A_\preventing,I_c,\mbox{centroid})$

\RETURN $a_\preventing$

\end{algorithmic}
\end{algorithm}
It receives as input \preventing's planning task $\Pi_\preventing$, the list of individual counterplanning landmarks $\mbox{CPList}$, and the current composite state $I_c$.
The algorithm ranks the counterplanning landmarks in $\mbox{CPList}$ to generate the set of candidate goals \textsc{ExtractCentroids} receives as input.
This ranking is computed using the following formula in the \textsc{Rank} function (line 2). We use \textsc{set} as a function that maps a list into a set.
\begin{equation}
\scriptsize
    {\cal G} = \Big\{ \langle G_i, w_i \rangle \mid w_i = \frac{\mbox{\textsc{count}}(G_i,\mbox{CPList})}{|\mbox{CPList}|} \hspace{2mm} \forall G_i \in  \textsc{set}(\mbox{CPList}) \Big\}
\end{equation}
Hence, counterplanning landmarks appearing in multiple goals will have a higher weight in ${\cal G}$, and therefore \textsc{ExtractCentroids} will prioritize those \emph{regions} of the state space.
Finally, the algorithm calls \textsc{GetFirstAction}, which returns the action that \preventing can execute from $I_c$ that minimizes the cost (steps) of achieving the centroid.
We compute only one action rather than a full plan to achieve the centroid since the plan might change in the next iteration of \algoritmoacp after observing a new action executed by \seeking.
This action will be the action returned by \textsc{anticipate}.

\begin{figure}
    \centering
    \includegraphics[scale=0.41]{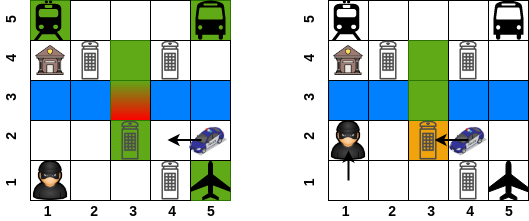}
    \caption[Behavior of \textsc{adicp} in the \textsc{police control} domain]{Behavior of \textsc{adicp} in a \textsc{police control} counterplanning task. Green cells depict individual counterplanning landmarks for each goal. Cells in red depict the centroid of the  counterplanning landmarks. The orange cell in the right image represents the goal selected by \preventing to block \seeking.}
    \label{fig:adicp_police}
\end{figure}

Let us exemplify how \algoritmoacp works in the \textsc{police control} counterplanning task depicted in Figure~\ref{fig:adicp_police}.
The start of the counterplanning episode (and the algorithm) is shown in the left image.
\seeking has not performed any action (no observation has been received), so all goals are equally likely.
The algorithm then extracts the list and set of counterplanning landmarks.
Since there is not common counterplanning landmark among the most likely goals (line 8), no valid counterplan can be produced yet, and the algorithm calls \textsc{anticipate} with the list of individual counterplanning landmarks $\mbox{CPList}$ (line 14).
The green cells depict the individual counterplanning landmarks.
The size of $\mbox{CPList}$ is $9$, i.e., there are $9$ individual counterplanning landmarks since some are repeated, i.e., they are common to several goals.
The counterplanning landmarks at each station appear only once in $\mbox{CPList}$.
Therefore, it assigns them a weight of $\frac{1}{9} = 0.11$.
On the other hand, the three counterplanning landmarks in the middle of the image (\textit{(not (free c3-2)}, \textit{(not (free c3-3))}, \textit{(not (free c3-4))}) appear twice in $\mbox{CPL}$, for \seeking escaping through the train and bus stations.
Therefore, their associated weight is $\frac{2}{9} = 0.22$.
This would be the set of weighted goals returned by the \textsc{Rank} function from the list of counterplanning landmarks:
\begin{equation*}
\scriptsize
\begin{split}
    {\cal G} = \{ 
    \langle (not (free c1\mbox{-}5)),0.11 \rangle, 
    \langle (not (free c5\mbox{-}5)),0.11 \rangle,\\
    \langle (not (free c5\mbox{-}1)),0.11 \rangle,
    \langle (not (free c3\mbox{-}4)),0.22 \rangle,\\
    \langle (not (free c3\mbox{-}3)),0.22 \rangle,
    \langle (not (free c3\mbox{-}2)),0.22 \rangle
    \}
\end{split}
\end{equation*}
Then, \textsc{anticipate} calls \textsc{ExtractCentroids}, which returns the next action that \preventing should execute: moving east from \textit{c5-2} to \textit{c4-2}.
This action makes the agent to be closer to most of the individual counterplanning landmarks.
After that, the composite state is updated with \seeking's next action $\pi^\prime_\seeking = \mbox{\textit{(move c1-1 c1-2)}}$ and the action prescribed by \textsc{anticipate}, $\pi^\prime_\preventing = \mbox{\textit{(move c5-2 c4-2)}}$.
This update ends the first iteration of \textsc{adicp}, and  \preventing gets a new observation. 

The second iteration is shown in the right image of Figure~\ref{fig:adicp_police}.
\seeking has moved north, so now the most likely goals are the terrorist escaping through the bus and train stations.
In this case there are counterplanning landmarks common to all the most likely goals ($\mbox{CPL} \neq \emptyset$), and thus the algorithm can start counterplanning (which is shared with \algoritmocp).
The first counterplanning landmark that \preventing can delete is also the closer to \seeking, \textit{(free c3-2)}, so its negation is set as $G_\preventing$.
Finally, a plan to achieve $G_\preventing$ is computed and the algorithm returns the counterplan $\pi_\preventing = \langle \textit{(move c4-2 c3-2)} \rangle$.
Neither this counterplan nor any other would have been existed if \preventing would not have moved towards the centroid of the counterplanning landmarks in the first place.
In other words, this is a counterplanning task that \algoritmocp cannot solve, but solvable for \algoritmoacp. 

On the other hand, if \seeking's goal would have been to escape through the airport, approaching the centroid would have been a bad decision, since the police patrol could have just moved south to block \seeking.
It is even possible to artificially generate counterplanning tasks where approaching the centroid of the counterplanning landmarks renders the counterplanning task unsolvable.
However, \algoritmoacp will prescribe the strategy that maximizes the chances of stopping \seeking on average, as we will see in the next section.

\section{Evaluation}
We compare four different algorithms: (1) \algoritmocp; (2) \algoritmoacp; (3) \textsc{random-adicp}, which is a variation of \algoritmoacp that executes random actions in the \textsc{anticipate} function; and (4) \textsc{random-goal-adicp}, which executes actions to achieve a random goal from ${\cal G}$ in the \textsc{anticipate} function.
The four algorithms can be instantiated with many different combination of planners, goal recognition, landmarks extraction, and planning centroid techniques.
However, for the sake of space we will run them with the same fixed configuration, leaving for future work the study of those different combinations.
For the \textsc{Recognize} function, we will use Ramírez and Geffner (\citeyear{DBLP:conf/aaai/RamirezG10}) probabilistic goal recognition approach.
We use the \texttt{seq-opt-lmcut} configuration of Fast Downward~\cite{DBLP:journals/jair/Helmert06} to optimally solve the compiled goal recognition problems using the \textsc{lmcut} admissible heuristic~\cite{DBLP:conf/aips/HelmertD09}.
We also use that optimal planner for the \textsc{Planner} function.
The (counter)plans in this case will be weak, and will only stop \seeking under some circumstances.
For the \textsc{ExtractCPL} and \textsc{ExtractListOfCPL} functions, we use (1) the Keyder, Richter and Helmert (\citeyear{DBLP:conf/ecai/KeyderRH10}) landmarks extraction algorithm as it is implemented in Fast Downward; and (2) the top-quality planner by Katz, Sohrabi and Udrea (\citeyear{DBLP:conf/aaai/KatzSU20}) to compute all the optimal plans.
Finally, we use Pozanco et al. (\citeyear{DBLP:conf/aips/PozancoEFB19}) greedy centroid computation to get the next action to execute in the \textsc{anticipate} function.

\begin{table*}[h!]
\centering
\small
\begin{tabular}{|l|l|c|c|c|c|c|}
\hline
\multicolumn{1}{|c}{Domain} & \multicolumn{1}{|c|}{Algorithm} & $\%E$ & $\%|\pi_\seeking|$ & $|\pi_\preventing|$ & $\%|\pi_\preventing|_{anticipate}$ & $t_{\cal C}$ \\ \hline
\multirow{4}{*}{ \textsc{police control} } & \textsc{dicp}  & $0.3 \pm 0.1$ & $0.8 \pm 0.2$ & $2.4 \pm 0.8$ & - & $\mathbf{37.3 \pm 10.8}$ \\ \cline{2-7}
& \textsc{adicp} & $\mathbf{0.5 \pm 0.2}$ & $\mathbf{0.6 \pm 0.2}$ & $7.2 \pm 2.8$ & $0.4 \pm 0.2$ & $39.8 \pm 17.1$\\ \cline{2-7}
& \textsc{random-adicp} & $0.3 \pm 0.2$ & $0.7 \pm 0.3$ & $8.1 \pm 3.4$ & $0.4 \pm 0.3$& $37.8 \pm 16.1$ \\ \cline{2-7}
& \textsc{random-goal-adicp} & $0.4 \pm 0.2$ & $0.7 \pm 0.1$ & $7.9 \pm 2.5$ & $0.5 \pm 0.2$& $37.5 \pm 15.9$\\ \hline \hline

\multirow{4}{*}{ \textsc{painted blocks-words} } & \textsc{dicp} & $0.3 \pm 0.2$ & $0.7 \pm 0.1$ & $4.6 \pm 1.8$& - & $\mathbf{46.1 \pm 12.2}$ \\ \cline{2-7}
& \textsc{adicp} & $\mathbf{0.7 \pm 0.1}$ & $\mathbf{0.6 \pm 0.2}$ & $5.0 \pm 2.3$ & $0.4 \pm 0.1$ & $48.3 \pm 15.5$ \\ \cline{2-7}
& \textsc{random-adicp} & $0.4 \pm 0.3$ & $0.7 \pm 0.2$ & $5.1 \pm 2.2$ & $0.4 \pm 0.1$ & $47.7 \pm 11.1$\\ \cline{2-7}
& \textsc{random-goal-adicp} & $0.5 \pm 0.2$ & $0.7 \pm 0.3$ & $4.9 \pm 2.1$ & $0.4 \pm 0.1$ & $49.2 \pm 13.0$ \\ \hline \hline

\multirow{4}{*}{ \textsc{rovers \& martians} } & \textsc{dicp} & $0.7 \pm 0.2$ & $0.5 \pm 0.3$ & $1.2 \pm 0.7$ & - & $\mathbf{52.3 \pm 22.1}$\\ \cline{2-7}
& \textsc{adicp} & $\mathbf{0.8 \pm 0.1}$ & $0.5 \pm 0.2$ & $1.2 \pm 0.7$ & $0.5 \pm 0.3$ & $61.7 \pm 20.4$ \\ \cline{2-7}
& \textsc{random-adicp} & $0.7 \pm 0.3$ & $0.5 \pm 0.3$ & $1.2 \pm 0.7$ & $0.5 \pm 0.3$ & $54.3 \pm 18.6$\\ \cline{2-7}
& \textsc{random-goal-adicp} & $\mathbf{0.8 \pm 0.1}$ & $0.5 \pm 0.2$ & $1.2 \pm 0.7$ & $0.5 \pm 0.3$ & $58.7 \pm 23.5$\\ \hline \hline

\multirow{4}{*}{ \textsc{human resources} } & \textsc{dicp}  & $0.7 \pm 0.2$ & $0.5 \pm 0.3$& $2.8 \pm 1.4$ & - & $\mathbf{41.6 \pm 20.9}$\\ \cline{2-7}
& \textsc{adicp} & $\mathbf{0.9 \pm 0.1}$ & $\mathbf{0.3 \pm 0.2}$ & $3.5 \pm 2.0$ & $0.3 \pm 0.1$ & $47.5 \pm 18.7$ \\ \cline{2-7}
& \textsc{random-adicp} & $0.7 \pm 0.3$ & $0.4 \pm 0.3$ & $3.1 \pm 1.7$ & $0.3 \pm 0.2$ & $42.3 \pm 20.1$\\ \cline{2-7}
& \textsc{random-goal-adicp} & $0.8 \pm 0.2$ & $0.4 \pm 0.1$& $3.2 \pm 2.1$ & $0.3 \pm 0.2$ & $46.3 \pm 20.5$\\ 
\hline
\end{tabular}
\caption{Comparison of the four counterplanning algorithms in the four considered domains. Numbers shown are all averages and standard deviations over the set of problems. Bold figures indicate the best performance in the given metric.}
\label{table_results}
\end{table*}

We compared the three algorithms in the following planning domains.
For each domain, we randomly generated 100 counterplanning tasks.

\textsc{\textbf{police control}}. Our running example domain.
    Maps are $10\times10$ grids with $25\%$ of obstacles and 10 randomly distributed booths.
    The set of candidate goals is $3$.

\textsc{\textbf{painted blocks-words}}. In this domain a robotic arm (\seeking) is trying to build some words using a set of blocks. It can stack and unstack blocks as long as their top is not painted.
    There is another agent (\preventing) that can paint the top part of clear blocks, i.e., blocks that do not have other blocks on top of them.
    The paint buckets 
    are randomly distributed over several connected rooms.
    \preventing is also randomly placed in one of these rooms.
    To paint a block, \preventing needs to have the paint and be in the room where \seeking is building the words.
    Problems in this domain contain $8$ blocks and $5$ rooms.
    Blocks are initially piled randomly into several towers.
    The set of candidate goals is $5$, i.e., \seeking might be building five words.
    The words \seeking needs to build range from $3$ to $6$ blocks (letters).
    
\textsc{\textbf{rovers \& martians}}. This is a game in which a robotic agent called rover (\seeking) has to conduct several experiments on Mars.
    It has to navigate to different locations, collect samples, analyze them and communicate the results.
    There is another agent, a martian (\preventing) that does not want intruders on its planet.
    It can destroy the rover's experiments by stealing the samples or interrupting its communications.
    While the rover can only move through visible locations (i.e., may need to apply a set of actions to go from A to B), the martian can move between any two points of the map with just one action.
    Problems in this domain contain 10 locations and 6 samples, in addition to different number of objectives and cameras.
    Both agents are randomly distributed over the map.
    The set of candidate goals is $6$, i.e., the rover might be trying to get/communicate the results of $6$ different experiments.
    
\textsc{\textbf{human resources}}. In this domain a company (\seeking) is trying to hire a set of recent graduates for a set of teams giving some budget limit over each team. 
    The company can execute some actions to increase the available budget of a team. 
    The graduates come from different universities, and have some preferences over the type of teams they would like to join, their salaries and their interviews' availability. 
    The company can establish contacts with universities in order to schedule interviews on different time slots and send offers to graduates. 
    There is another company (\preventing) hiring, also interested in these graduates.
    It can schedule interviews with the graduates in the same time slots, as well as thwarting the contacts \seeking can try to achieve with universities.
    Problems in this domain contain 10 universities, 50 graduates, 4 teams and 5 time slots in which the interviews can be scheduled.
    Teams' budgets and salaries are discretized in 10 consecutive bins.
    The set of candidate goals is $5$, i.e., $5$ different graduates that \seeking might be trying to hire. 

For all the counterplanning tasks, we randomly select one goal $G_i \in {\cal G}_\seeking$ and set it as \seeking's true goal.
Then we compute an optimal plan to achieve it ($\pi_\seeking$) using the \textsc{Planner} function.
We measure the following quality and performance metrics:
\begin{itemize}
    \item $\%E$: $1$ if \seeking is stopped, $0$ otherwise. To compute this number, we jointly execute from the beginning ($I_c = I_\seeking \cup I_\preventing$) $\pi_\seeking$ and $\pi_\preventing$ as returned by the algorithms.
    If the joint execution of both plans from $I_c$ does not allow \seeking to achieve its goal, $E=1$. 
    
    \item $\%|\pi_\seeking|$: ratio of $\pi_\seeking$ that \seeking can execute until it is blocked by \preventing. Lower numbers are better, and $1$ means that \seeking has not been blocked. 
    
     \item $|\pi_\preventing|$: counterplan length (cost) for \preventing. 
     
     \item $\%|\pi_\preventing|_{a}$: ratio of actions from $\pi_\preventing$ that are part of $\pi^\prime_\preventing$, i.e., that were generated before the counterplan was computed. 
     
     \item $t_{\cal C}$: time to return a solution for a counterplanning task.
     We compute this time as the average over all the iterations of the algorithms.
\end{itemize}
For all the metrics (except for $\%E$) we only report the numbers in those cases where the given algorithm could find a valid counterplan, i.e., we do not report those metrics when the counterplanning task is unsolvable or when the generated (weak) counterplan is not valid.
The experiments were run on Intel Core i5-8400 CPU 2.80GHz machines with a time limit of 600s and a memory limit of 8GB per algorithm iteration.
Domains and problems are available upon request.

\subsection{Results}

Table~\ref{table_results} summarizes the results of our evaluation.
As we can see, \algoritmoacp dominates the other algorithms in all domains without much overhead in terms of computation time.
It is able to return valid counterplans ($\%E$) in more counterplanning tasks, also stopping \seeking before, i.e., allowing it to execute less part of its plan ($\%|\pi_\seeking|$).
\textsc{random-goal-adicp} achieves comparable results in domains like \textsc{rovers \& martians}.
However their performance difference increases in others such as \textsc{painted blocks-words}, highlighting the fact that \textsc{adicp}'s strategy of approaching the centroid of the candidate goals is better than just starting to move towards one of the candidate goals.
A key observation is that even \textsc{random-adicp} tends to behave better than \textsc{dicp}. Thus, in most counterplanning tasks, randomly moving until a counterplan is found is usually better than doing nothing  as \textsc{dicp} does.

The average length of the plans of the three algorithms with an anticipatory component tends to be greater than those of \algoritmocp.
This is specially noticeable in \textsc{police control}, where the counterplans generated by \algoritmoacp triple the length of those generated by \algoritmocp.
First, these algorithms solve more problems than \algoritmocp. And, second, the problems that could not be solved by \algoritmocp are the harder tasks, i.e., problems where $G_\seeking$'s counterplanning landmarks are further away from \preventing's initial state.
The $\%$ of \preventing's plan with anticipatory actions differs across domains and algorithms, but it represents around $30$-$50\%$ of the plan length.

\section{Related Work}
Different approaches solve problems where agents try to prevent opponents from achieving their goals~\cite{DBLP:journals/ai/Carbonell81,DBLP:conf/iaw/Rowe03}. Stackelberg games~\cite{stackelberg1952theory} are among the most successful and prolific ones~\cite{DBLP:books/daglib/0040483}, where
the \textit{leader} (defender) moves first, followed by the \textit{follower} (attacker). 
A solution is an equilibrium pair that minimizes the defender's loss given an optimal attacker (opponent) play.
Stackelberg planning~\cite{DBLP:conf/aaai/SpeicherS00K18,DBLP:conf/aaai/TorralbaSKS021} has been recently introduced as a way of computing solutions to these games using automated planning.
However unlike ours, these works assume (1) agents act sequentially, i.e., one after the other; and (2) the opponent's goal is known a priori.
In adversarial scenarios where agents execute concurrently, Jensen and Veloso (\citeyear{DBLP:journals/jair/JensenV00}) present two algorithms: optimistic and strong cyclic adversarial planning. 
These algorithms return universal plans, which can be seen as policies that associate the best action to apply given a state for achieving the goal.
However, they also assume the opponent's goal is known, while we use planning-based goal recognition to infer it from a set of candidate goals.

In this paper, agents can reason about the goal they should pursue based on the environment~\cite{DBLP:conf/aaai/MolineauxKA10a}.
The \preventing agent changes its goal at each step, approaching first the centroid of the counterplanning landmarks (which changes over time), to later change its behavior when it can compute a counterplan.
We also provide agents with proactive behavior that allows them to start planning before their actual goal appears (as a reaction to the deduction of the opponent's goal).
This kind of anticipatory planning has been explored in the past in the context of single agent planning~\cite{DBLP:conf/aips/BurnsBRYD12,DBLP:journals/aicom/FuentetajaBR18} in settings where goals arrive dynamically and the planner must generate plans to start achieving them before they appear.

\section{Conclusions and Future Work}
In this paper we have shown that there are some counterplanning tasks that are unsolvable due to the preventing agent inactivity until it is able to infer its opponent's goal.
To overcome this limitation, we have introduced \algoritmoacp, a new algorithm that combines a set of planning techniques (goal recognition, landmarks, and  centroids) to yield proactive plans that greatly increase the chances of blocking an opponent in different adversarial planning domains.

In future work, we would like to study in more depth the impact of some key aspects in the performance of anticipation. For instance,
deceptive seeking agents~\cite{DBLP:conf/aaai/KulkarniSK19} that actively try to mislead the preventer, as well as problem distribution (placement of initial state and goals) or domain definition (actions costs on both sides).

\newpage
\clearpage

\bibliographystyle{aaai}
\bibliography{references}

\end{document}